\documentclass[conference]{IEEEtran}
\IEEEoverridecommandlockouts
\usepackage{cite}
\usepackage{amsmath,amssymb,amsfonts}
\usepackage{graphicx}
\usepackage{textcomp}
\usepackage{xcolor}

\usepackage{algorithm}
\usepackage{algpseudocode}

\usepackage{multirow}

\usepackage{orcidlink}

\def\BibTeX{{\rm B\kern-.05em{\sc i\kern-.025em b}\kern-.08em
    T\kern-.1667em\lower.7ex\hbox{E}\kern-.125emX}}
\begin{document}

\title{Uncertainty-Aware Federated Learning for Infant Movement Analysis}% Using Skeletal Motion Data}

\author{\IEEEauthorblockN{Edmond S. L. Ho}
\IEEEauthorblockA{\textit{School of Computing Science} \\
\textit{University of Glasgow}\\
Glasgow, Scotland, United Kingdom \\
Shu-Lim.Ho@glasgow.ac.uk, \orcidlink{0000-0001-5862-106X}}
}

\maketitle

\begin{abstract}
Infant movement analysis provides valuable biomarkers for the early identification of neurodevelopmental disorders. Recent advances in deep learning have enabled automated analysis of infant movements from video-derived skeletal representations, achieving performance comparable to expert assessment for tasks such as General Movement Assessment (GMA). However, most existing approaches rely on centralized training, requiring data from multiple institutions to be collected and stored at a single site. Such assumptions are often impractical in clinical settings due to privacy, governance, and data-sharing constraints. To address these challenges, we present, to the best of our knowledge, the first federated learning framework for automated infant movement analysis and General Movement Assessment using skeletal motion data. As a clinically relevant use case, the proposed framework is evaluated on fidgety movement classification. To quantify model confidence, Monte Carlo (MC) Dropout is employed to estimate predictive uncertainty during inference. Building upon this, we propose an Uncertainty-Aware Federated Averaging (UA-FedAvg) strategy that incorporates predictive entropy derived from MC-Dropout into the federated aggregation process, enabling client contributions to be adjusted according to their predictive uncertainty. Experiments were conducted using a cross-subject evaluation protocol under a three-client federated learning setting. Results demonstrate that federated learning substantially improves classification performance compared with independently trained local models while achieving performance approaching that of centralized training. Furthermore, UA-FedAvg and its variant incorporating validation loss generally outperform conventional FedAvg across the evaluated data-split configurations. These findings suggest that predictive uncertainty may provide useful information for guiding federated aggregation in infant movement analysis.

\end{abstract}


\section{Introduction}
General Movement Assessment (GMA) identifies several distinct abnormal movement patterns across two developmental phases. %During the writhing movement period, three categories of abnormal GMs have been described \cite{einspieler2005prechtl,EINSPIELER199747}: \textit{poor-repertoire} (PR), characterised by monotonous movement sequences with reduced variability in intensity, speed, and amplitude; \textit{cramped-synchronised} (CS), where limb and trunk muscles contract and relax almost simultaneously, resulting in rigid movement patterns; and \textit{chaotic}, which consists of abrupt, large-amplitude, and high-speed movements that are rarely observed beyond the preterm period. 
During the fidgety movement period, spontaneous movements are categorized as normal (FM+), \textit{absent} (FM-), or \textit{abnormal} \cite{einspieler2005prechtl,EINSPIELER199747}. Among these patterns, persistent Cramped-Synchronized (CS) during the writhing period and FM- during the fidgety period have consistently been reported as the strongest predictors of later Cerebral Palsy (CP)~\cite{einspieler2005prechtl}. Consequently, most automated GMA systems focus on binary or multi-class classification problems derived from these categories, with the majority of studies targeting the discrimination between FM+ and FM-. %\cite{irshad2020ai,silva2021future,deng2025systematic}.

Despite the rapid progress in automated GMA, most existing studies adopt a centralized learning paradigm in which all training data are pooled into a single repository. In clinical practice, however, sharing infant movement recordings across institutions is often restricted by privacy, ethical, and governance requirements~\cite{SPITTLE2025103379}. In addition, data collected at different sites may exhibit distributional differences arising from variations in demographics, recording protocols, and acquisition settings. These challenges motivate the use of federated learning, which enables multiple institutions to collaboratively train a shared model without exchanging raw data. By keeping data locally at each site and aggregating only model updates, federated learning offers a practical approach for leveraging distributed infant movement datasets while accommodating inter-site heterogeneity. However, differences in local data distributions may result in varying levels of model confidence across clients, introducing challenges that remain largely unexplored in automated GMA.

Federated Learning (FL) was introduced by McMahan et al.~\cite{FedAvg} as a distributed learning paradigm that enables multiple clients to collaboratively train a shared model without exchanging raw data. %Instead of centrally collecting all training samples, each client performs local optimisation using its private dataset and transmits only model updates to a central server for aggregation. 
The most widely adopted federated optimisation algorithm is Federated Averaging (FedAvg)~\cite{FedAvg}, in which the global model is obtained through a weighted average of the client model parameters according to the number of local training samples. FL has attracted considerable attention in healthcare due to increasing concerns regarding patient privacy, regulatory constraints, and the logistical challenges of sharing sensitive medical data across institutions~\cite{rieke2020future,sheller2020federated}. More recently, researchers have begun exploring uncertainty-aware federated learning approaches. Unlike conventional FL methods that primarily rely on dataset size when weighting client updates, uncertainty-aware approaches attempt to assess the reliability of local models and incorporate this information into the optimisation process. Existing studies have investigated uncertainty estimation, ensemble-based approaches, and Monte Carlo (MC) Dropout techniques to quantify prediction confidence under heterogeneous client distributions~\cite{zhang2025uncertainty}. %Other work has demonstrated that uncertainty information can be used to improve client selection, client weighting, and aggregation robustness under highly non-IID conditions~\cite{chen2025ufl}. 
Bhatt et al.~\cite{Bhatt:ACML2023} proposed a Bayesian federated learning approach which aggregates client predictive distributions through a distillation framework. These findings suggest that uncertainty estimation may provide complementary information beyond conventional performance metrics when determining the contribution of individual clients to the global model. Although federated learning has been widely investigated in healthcare applications, limited attention has been given to automated GMA.

In this work, we adopt a federated learning framework for infant movement classification and incorporate predictive uncertainty into model aggregation. Specifically, MC-Dropout is used to estimate client-level predictive uncertainty, which is subsequently used to modulate each client's contribution to the global model update. The proposed approach seeks to account for differences in model confidence across clients while preserving the data-locality benefits of federated learning. Unlike entropy-based client-selection methods that rely on dataset statistics or label-distribution entropy, our method derives uncertainty directly from model predictions and incorporates it into the aggregation process.

%To the best of our knowledge, uncertainty-aware federated learning has not been investigated for automated General Movement Assessment. This work therefore examines whether uncertainty-informed aggregation can improve the reliability and generalisability of federated infant-movement classification models operating across heterogeneous client datasets. Code and experiment scripts are publicly available on GitHub.

To the best of our knowledge, this represents the first application of federated learning to automated General Movement Assessment (GMA), and the first study to investigate uncertainty-aware federated learning in this setting. We therefore evaluate whether predictive-uncertainty-informed aggregation can improve federated infant movement classification under heterogeneous client data distributions. The source code and experimental scripts are publicly available on GitHub.\footnote{\url{https://github.com/edmondslho/UA-FedAvg}}

%In this work, we address this limitation by integrating uncertainty estimation into the federated learning process, allowing client contributions to be weighted according to their predictive confidence during model aggregation. Inspired by recent uncertainty-aware federated learning approaches, we utilize MC-Dropout-derived predictive entropy as a client-level reliability indicator during federated aggregation. 

\section{Related Work}
\subsection{Automated General Movement Assessment}  The development of automated GMA systems has progressed considerably over the past decade, spanning a wide range of sensing modalities, feature representations, and classification strategies. % \cite{irshad2020ai,silva2021future,deng2025systematic}. 
%Early approaches primarily employed conventional machine learning pipelines, in which handcrafted motion descriptors, such as optical flow statistics, limb displacement measures, and frequency-domain features, were extracted from video recordings and subsequently classified using machine learning algorithms including support vector machines and random forests \cite{adde:2009,adde2010early,Stahl:TNSRE2012,rahmati2016frequency}. Although these methods demonstrated the feasibility of computer-assisted GMA, their performance depended heavily on feature engineering and they were often unable to capture the complex spatio-temporal movement patterns underlying the clinical Gestalt assessment performed by expert assessors. 
Recent developments in computer vision have shifted the focus towards pose-based representations. Most modern automated GMA systems first extract skeletal joint coordinates from video recordings using pose-estimation frameworks such as ViTPose \cite{xu2022vitpose}, or infant-specific pose estimation models \cite{Chambers:TNSRE2020,Pmi-GMA}. Following pose extraction, existing approaches can broadly be categorized into two groups. The first group continues to rely on handcrafted kinematic representations, deriving features such as joint angles, velocities, movement ranges, and their statistical summaries before classification using traditional machine learning or shallow deep-learning models \cite{McCay:TNSRE2022,McCay:DeepBaby,doroniewicz2020writhing,McCay:EMBC2019,McCay:BHI2021}. The second group adopts end-to-end learning strategies, where raw skeletal sequences are directly processed by deep neural networks, including convolutional neural networks (CNNs) and graph convolutional networks (GCNs), to jointly learn feature representations and diagnostic decision boundaries \cite{kulvicius2025deep,Nguyen-Thai:JBHI2021,Sakkos:Access2021,luo2025udf,Luo:MICCAI2022,Zhu:EMBS2021,Pmi-GMA,pellano2026towards}. %Within this category, skeleton-based action-recognition architectures such as ST-GCN~\cite{STGCN} and CTR-GCN \cite{CTRGCN} have been particularly influential. 

\subsection{Federated Learning for Healthcare AI} By allowing hospitals and healthcare providers to collaboratively train machine learning models without directly exchanging patient data, FL has been successfully applied to a variety of healthcare problems, including medical image segmentation, disease diagnosis, patient risk prediction, and physiological signal analysis~\cite{sheller2020federated}. %,dayan2021federated}. 
These studies have demonstrated that federated models can often achieve performance comparable to centrally trained models while preserving data ownership and confidentiality. Despite these advantages, practical FL deployments face significant challenges arising from data heterogeneity. Data collected by different institutions frequently exhibit non-independent and non-identically distributed (non-IID) characteristics due to variations in patient demographics, acquisition protocols, sensor configurations, and disease prevalence~\cite{kairouz2021advances}. 
Such heterogeneity can lead to client drift, slower convergence, and degraded global model performance when standard FedAvg aggregation is employed~\cite{li2020fedprox,SCAFFOLD}. To address these issues, numerous aggregation strategies have been proposed, including FedProx~\cite{li2020fedprox}, which constrains local updates to remain close to the global model, SCAFFOLD~\cite{SCAFFOLD}, which introduces control variates to reduce client drift, and adaptive optimisation methods such as FedAdam and FedYogi~\cite{reddi2021adaptive}, which incorporate server-side adaptive learning rates.

\section{Methodology}
The overview of the proposed framework is illustrated in Figure~\ref{fig:overview}. In this section, we will first formulate the downstream task, i.e. automated General Movement Assessment (GMA) in Section~\ref{sec:problem}. Next, the dataset used in this study will be presented in Section~\ref{sec:dataset}. For the federated learning pipeline, the details are available in Section~\ref{sec:FL}. %Finally, different settings explored in this study will be shared in Section [y]. [Uncertainty?] 

\subsection{Problem Formulation} \label{sec:problem}
In General Movement Assessment (GMA), the main focus is to analyze the body movements of infants, for example, detecting any fidgety movements (FMs), to indicate any atypical movements which can potentially be used for predicting any underlying neurological disorders, such as early prediction of Cerebral Palsy (CP). %In the automated setting, researchers have been taking advantage of using advanced Computer Vision models to extract discriminative movement features from the RGB videos to train machine learning models (i.e. classifiers) to automate GMA. 
The mainstream of recent automated GMA~\cite{McCay:EMBC2019,McCay:TNSRE2022,Passmore:2024,kulvicius2025deep} focus on using 2D skeletal pose sequences extracted from video using pose estimation models, such as ViTPose~\cite{xu2022vitpose}, to train machine learning models to classify the presence or absence of FMs. Using such a compact motion representation can facilitate the training of models due to the low dimensionality and removal of irrelevant information such as the variations introduced by lighting and clothing when compared with classifying the raw RGB videos.

In this study, we follow this stream to focus on training a binary classifier $f: M \to \{0, 1\}$ to map the input skeletal motion $M \in \mathbb{R}^{F\times d}$, where $F$ and $d$ denote the number of frames and feature dimension, respectively, to a binary label indicating the presence or absence of FMs.

\subsection{Dataset} \label{sec:dataset}
The skeletal motion data used in this study were obtained from the publicly available multimodal infant movement dataset introduced by Kulvicius et al.~\cite{kulvicius2025deep}. The dataset comprises 2D skeletal motion data extracted from RGB videos, together with inertial measurement unit (IMU) recordings and pressure mattress sensor data collected from 45 infants (23 females) with typical development (TD). Data acquisition was performed between 4 and 16 weeks post-term age. In the present study, only the 2D skeletal motion modality was considered. Following the protocol adopted in~\cite{kulvicius2025deep}, each video recording was segmented into non-overlapping 5-second clips, corresponding to the minimum duration required by certified General Movement (GM) Assessors to evaluate movement quality. All clips were independently annotated as either the presence (FM+) or absence (FM-) of fidgety movements by two senior GM Assessors with more than 20 years of clinical experience. In total, 1,683 clips were available for analysis, comprising 943 FM+ clips and 740 FM- clips. 

The 2D skeletal pose sequences were extracted from the RGB videos using ViTPose~\cite{xu2022vitpose}. While ViTPose estimates 17 body keypoints by default, the two ear keypoints were removed by the dataset authors as they were considered redundant for characterising head orientation and movement patterns~\cite{kulvicius2025deep}. To better capture movement dynamics, the dataset additionally provides the temporal velocities of the $x$- and $y$-coordinates for each joint. Consequently, each 5-second video clip is represented as a skeletal motion sequence $M \in \mathbb{R}^{250 \times 60}$, where 250 corresponds to the number of frames recorded at 50 frames per second (fps) over a 5-second interval. The feature dimension of 60 is derived from the 15 retained body joints, with each joint represented by a four-dimensional feature vector consisting of the spatial coordinates $(x,y)$ and their corresponding velocities $(v_x,v_y)$. Prior to model training, the skeletal data were pre-processed to remove outliers and standardized to improve numerical stability and facilitate network optimisation. Readers are referred to~\cite{kulvicius2025deep} for a detailed description of the data acquisition process, pose extraction pipeline, and pre-processing procedures.

\section{Federated Learning Pipeline} \label{sec:FL}
\begin{figure*}[htbp]
    \centering
    \includegraphics[width=0.7\linewidth]{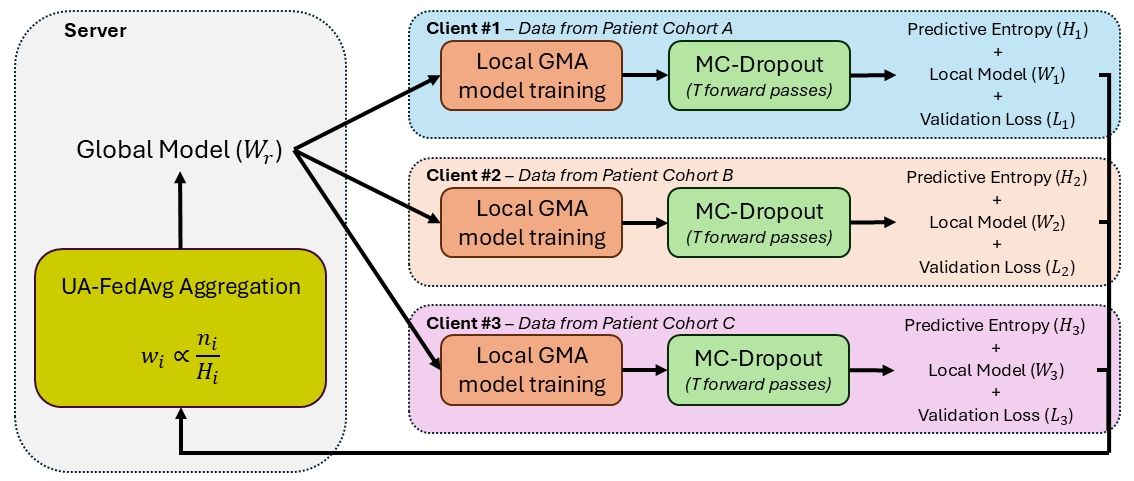}
    \caption{The overview of the proposed federated learning framework for multi-client General Movement Assessment model training.}
    \label{fig:overview}
\end{figure*}

In this study, we simulate the situation in which the clients (e.g. hospitals) have no access to the data from other clients to train their local automated GMA model, and each client has a distinct dataset which was collected from their own patient cohort. Given the size of the dataset~\cite{kulvicius2025deep} (collected from 45 subjects), we divided the data into 3 subsets according to the subject IDs to maintain a good balance between the number of clients and the amount of data for training a local model. The pipeline is illustrated in Figure~\ref{fig:overview}.

\subsection{Client - Automated GMA model}
\begin{figure}[htbp]
    \centering%
    \includegraphics[width=1\linewidth]{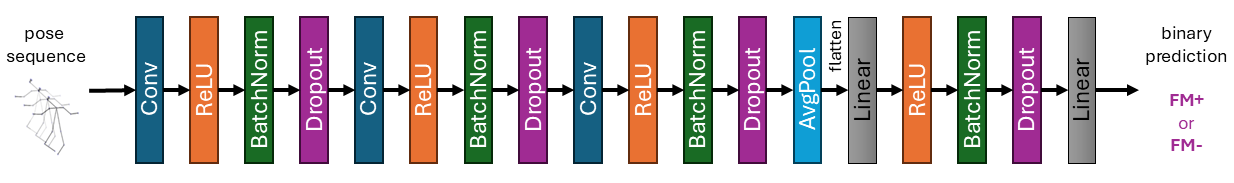}
    \caption{The architecture of the CNN for skeletal motion classification trained at each client.}
    \label{fig:CNN_arch}
\end{figure}

As stated in Section~\ref{sec:problem}, a classifier will be trained to map the input motion $M$ into a binary label to indicate the presence (FM+) or absence (FM-) of fidgety movement. Here, we adapt the CNN-based classification model proposed in \cite{kulvicius2025deep} since the model achieved high accuracy on the dataset. We used the optimal hyperparameters stated in \cite{kulvicius2025deep} in all experiments. The network architecture is illustrated in Figure~\ref{fig:CNN_arch}.

\subsection{Client - Uncertainty Estimation}
Although previous work~\cite{kulvicius2025deep} demonstrated strong classification performance, conventional performance metrics such as accuracy, sensitivity, and specificity do not provide information regarding the confidence of the model predictions. As highlighted by Gal and Ghahramani~\cite{gal2016dropout}, modern deep neural networks can produce highly confident predictions even when presented with previously unseen or out-of-distribution samples. This limitation is particularly relevant in federated learning environments, where data distributions may differ substantially across participating clients due to variations in subject cohorts, recording conditions, and demographic characteristics. 

To address this issue, uncertainty quantification was incorporated into the proposed federated learning framework alongside conventional classification metrics. Specifically, predictive uncertainty was estimated using Monte Carlo (MC) Dropout. Predictive uncertainty reflects the overall uncertainty of the model's predictions and captures both uncertainty arising from limitations in the learned model and uncertainty associated with ambiguous or noisy observations. The resulting uncertainty estimates were subsequently used to guide client weighting during federated aggregation, with lower predictive uncertainty indicating higher model confidence.

\subsubsection{Predictive Uncertainty Estimation Using MC-Dropout}
Predictive uncertainty was estimated using Monte Carlo Dropout (MC-Dropout)~\cite{gal2016dropout}. During inference, dropout layers remained active and multiple stochastic forward passes were performed through the trained network. Given an input sample $x$, the model prediction obtained from the $t^{th}$ stochastic forward pass is denoted by
\begin{equation}
p_t(y|x), \quad t = 1,\ldots,T,
\end{equation}
where $T$ represents the number of Monte Carlo samples. The predictive mean probability is computed as
\begin{equation} \label{eq:PP}
\bar{p}(y|x) = \frac{1}{T} \sum_{t=1}^{T} p_t(y|x). 
\end{equation}

% predictive variance, which serves as a direct estimate of epistemic uncertainty, is calculated as
%\begin{equation}
%\mathrm{Var}[p(y|x)] = \frac{1}{T} \sum_{t=1}^{T} \left( p_t(y|x) - %\bar{p}(y|x) \right)^2.
%\end{equation}

%A high predictive variance indicates substantial disagreement among the sampled models and therefore reflects greater epistemic uncertainty. Conversely, a low predictive variance suggests that the model consistently produces similar predictions and is therefore more certain about its decision.

The predictive variance across Monte Carlo samples is calculated as 
\begin{equation}
\mathrm{Var}[p(y|x)] = \frac{1}{T} \sum_{t=1}^{T} \left( p_t(y|x) - \bar{p}(y|x) \right)^2.
\end{equation}
which quantifies the variability of the model predictions obtained from repeated stochastic forward passes. A high predictive variance indicates substantial disagreement among the sampled predictions and therefore reflects greater predictive uncertainty. Conversely, a low predictive variance suggests that the model consistently produces similar predictions and is therefore more confident in its decision.

\subsubsection{Predictive Uncertainty of Client Models}
%While predictive variance provides a direct estimate of epistemic uncertainty, it was used to quantify epistemic uncertainty at the sample level. On the other hand, predictive entropy was adopted as a client-level reliability measure for federated aggregation. In the context of MC-Dropout, the stochastic forward passes approximate samples drawn from the posterior distribution of the model parameters~\cite{gal2016dropout}. 

%When the sampled models produce substantially different predictions for the same input, the predictive variance increases, indicating that the model is uncertain about its prediction due to insufficient knowledge acquired during training. Such behaviour is characteristic of high epistemic uncertainty. From a probabilistic perspective, elevated epistemic uncertainty generally results in a broader predictive distribution and consequently higher predictive entropy. In contrast, when the sampled models consistently produce similar predictions, the predictive distribution becomes more concentrated, resulting in lower entropy and indicating a higher degree of confidence. Therefore, predictive entropy can be interpreted as a surrogate measure of the confidence and reliability of a local client model.

While predictive variance characterizes uncertainty at the sample level, predictive entropy was adopted as a client-level uncertainty measure for federated aggregation. In the context of MC-Dropout, repeated stochastic forward passes provide multiple predictions for the same input, from which a predictive distribution can be estimated~\cite{gal2016dropout}. 

Predictive entropy captures the uncertainty of this predictive distribution. A more concentrated predictive distribution results in lower entropy and indicates greater model confidence, whereas a more dispersed distribution results in higher entropy and reflects greater uncertainty in the model's predictions. %In this study, the predictive entropy was averaged across the local validation set to obtain a client-level uncertainty score. This score was subsequently incorporated into the federated aggregation process, with lower predictive entropy corresponding to larger aggregation weights.

%Within the proposed federated learning framework, each client is trained on a distinct local dataset that may differ in size, diversity, and underlying data distribution. Clients with smaller or less representative datasets are expected to exhibit greater epistemic uncertainty, as their local models have access to less information about the overall population. Consequently, updates generated by such clients may be less reliable and may contribute to model drift during the aggregation process. Conversely, clients with lower uncertainty are more likely to have learned feature representations that generalise well to unseen samples.

Within the proposed federated learning framework, each client is trained on a distinct local dataset that may differ in size, diversity, and underlying data distribution. As a result, the predictive uncertainty estimates obtained from MC-Dropout may vary across clients. In this work, predictive entropy is used as a client-level uncertainty score to modulate the contribution of local model updates during federated aggregation, with lower predictive entropy corresponding to larger aggregation weights. In particular, predictive entropy was computed as a summary measure of predictive uncertainty:
\begin{equation} \label{eq:PE}
H(x) = -\bar{p}(y|x)\log \bar{p}(y|x) - (1-\bar{p}(y|x)) \log(1-\bar{p}(y|x)). \end{equation}

%The mean predictive entropy across the local validation dataset was subsequently used to estimate the overall confidence of each client model. Within the proposed uncertainty-aware federated aggregation strategy, clients exhibiting lower predictive entropy were assigned greater influence during model aggregation, reflecting a higher degree of confidence in their learned representations. The algorithm of the client-side uncertainty estimation is summarized in Algorithm~\ref{alg:mcdropout}.

The mean predictive entropy across the local validation dataset was computed to obtain a client-level uncertainty score, which was subsequently used during federated aggregation. The algorithm for client-side uncertainty estimation is summarized in Algorithm~\ref{alg:mcdropout}.

\begin{algorithm}[htb] 
    \caption{Client-Side MC-Dropout Uncertainty Estimation} 
    \label{alg:mcdropout} 
    \begin{algorithmic}[1] 
        \Require Validation samples $V$, number of validation samples $N$, trained model $W$, MC samples $T$ 
        \ForAll{sample $x \in V$} 
            \For{$t=1$ to $T$} 
                \State Enable dropout 
                \State Compute prediction $p_t(x)$ 
            \EndFor 
            \State Compute predictive mean using Eq.~\ref{eq:PP}%\[ \bar{p}(x) = \frac{1}{T} \sum_{t=1}^{T} p_t(x) \] 
            \State Compute predictive entropy using Eq.~\ref{eq:PE}%\[ H(x) = -\bar{p}(x)\log \bar{p}(x) -(1-\bar{p}(x)) \log(1-\bar{p}(x)) \] 
        \EndFor 
        \State Compute client uncertainty \[ H_i = \frac{1}{N} \sum_{x \in V} H(x) \] 
        \State \Return $H_i$ 
    \end{algorithmic} 
\end{algorithm}

\subsection{Server - Federated Aggregation}
A federated learning framework based on Flower~\cite{beutel2020flower} was employed to enable collaborative model training across three distributed clients without sharing raw infant movement data. Each client maintained its own local dataset and trained an identical CNN model initialized from the current global model. The steps are summarized in Algorithm~\ref{alg:ufl}.

\begin{algorithm}[htb]
\caption{Uncertainty-Aware Federated Learning} \label{alg:ufl}
\begin{algorithmic}[1] 
\Require Number of clients $K$, communication rounds $R$, local epochs $E$ 
\State Initialize global model parameters $W_0$
\State Initialize best validation loss $L_{\text{best}} \gets \infty$
\For{$r = 1$ to $R$} \ForAll{clients $i \in \{1,\dots,K\}$ \textbf{in parallel}} 
\State Receive global model $W_r$ 
\State Train local CNN on training data for $E$ epochs 
\State Obtain updated model $W_i$ 
\State Evaluate $W_i$ on local validation set 
\State Compute validation loss $L_i$ 
\State Perform MC-Dropout with $T$ stochastic passes 
\State Compute mean predictive entropy $H_i$ 
\State Send $\{W_i, L_i, H_i, n_i\}$ to server 
\EndFor 
\State Compute uncertainty-aware client weights using Eq.~\ref{eq:PEAgg}%: \[ \alpha_i = \frac{\frac{n_i}{1+H_i}} {\sum_{j=1}^{K}\frac{n_j}{1+H_j}} \] 
\State Aggregate client models using Eq.~\ref{eq:AggMdl} %: \[ W_{r+1} = \sum_{i=1}^{K} \alpha_i W_i \] 
\State Update global model and send it to clients
\State Aggregate validation loss using Eq.~\ref{eq:LGlobal}%: \[ L_{\text{global}} = \frac{\sum_{i=1}^{K} n_i L_i} {\sum_{i=1}^{K} n_i} \] 
\If{$L_{\text{global}} < L_{\text{best}}$} 
\State $L_{\text{best}} \gets L_{\text{global}}$ \State Save global model $W_{r+1}$ \EndIf 
\EndFor 
\State \Return Best global model 
\end{algorithmic} 
\end{algorithm}

At the start of each federated round $r$, the server distributed the global model parameters $W_r$ to all participating clients. Each client then performed local training for $E$ epochs using its private dataset and returned the updated model parameters together with validation metrics to the server.

The server aggregated the received model updates to obtain the updated global model $W_{r+1}$.

For the aggregation, two approaches were used:

\subsubsection{Standard FedAvg}
A standard FedAvg~\cite{FedAvg} was used in the study, with the global model updated by
\begin{equation}
    W_{r+1} = \sum^{K}_{i=1}\frac{n_i}{\sum_{j}n_j}W_i
\end{equation}
where $K$ is the number of clients (3 in our study), $n_i$ is the number of training samples and $W_i$ is the locally trained model parameters at client $i$.

\subsubsection{Uncertainty-Aware FedAvg (UA-FedAvg)}
We also explored using the predictive entropy to inform the aggregation:
\begin{equation} \label{eq:PEAgg}
    %w_i=\frac{\frac{n_i}{1+H_i}}{\sum_{j}\frac{n_j}{1+H_j}}
    w_i= \frac{ \frac{n_i}{H_i+\epsilon} } { \sum_j \frac{n_j}{H_j+\epsilon} }
\end{equation}
where $H_i$ represents the mean predictive entropy estimated using MC-Dropout on the validation dataset of client $i$, and $\epsilon$ is a small constant introduced for numerical stability. We further proposed a variant, \textit{UA-FedAvg w/ vLoss}, to incorporate both predictive uncertainty and model performance when computing the aggregation weight of client $i$:
\begin{equation} 
w_i= \frac{ \frac{n_i} {1+\alpha(L_i+H_i)} } { \sum_j \frac{n_j} {1+\alpha(L_j+H_j)} }, 
\end{equation}
where $L_i$ denotes the validation loss, and $\alpha$ is a scaling factor controlling the influence of the uncertainty and validation-loss terms. In this study, $\alpha$ was empirically set to 0.25.

The global model is then updated by
\begin{equation} \label{eq:AggMdl}
    W_{r+1} = \sum^{K}_{i=1}w_iW_i
\end{equation}

%By this, the weighting strategy assigns greater influence to client updates exhibiting lower predictive uncertainty. The underlying assumption is that local models with lower predictive entropy are more reliable estimators of the underlying infant movement patterns and therefore should exert greater influence on the global model update. By incorporating uncertainty information into the aggregation process, the proposed framework aims to mitigate the impact of noisy, poorly generalised, or highly uncertain local models, thereby improving the robustness and stability of federated optimisation in heterogeneous multi-client environments.

By this, the weighting strategy assigns greater influence to client updates exhibiting lower predictive uncertainty. The approach is motivated by the hypothesis that predictive uncertainty may provide a useful proxy for client confidence when aggregating local model updates. The effectiveness of this uncertainty-aware aggregation strategy is subsequently evaluated through the experimental studies presented in Section~\ref{sec:res}.

\subsection{Global Model Selection}
To avoid selecting the final model solely based on the last communication round, each client evaluated the received global model on its local validation set after every federated round. The server aggregated the validation losses using a weighted average:
\begin{equation} \label{eq:LGlobal}
    L_{global} = \frac{\sum^{K}_{i=1}n_iL_i}{\sum^{K}_{i=1}n_i}.
\end{equation}
%where $L_i$ denotes the validation loss of client $i$.

The global model corresponding to the minimum aggregated validation loss observed during training was retained as the final model and stored as the best global checkpoint.

\section{Experiments}
\subsection{Experimental Settings}
We adopted the same 9-fold cross-subject validation protocol proposed by Kulvicius et al.~\cite{kulvicius2025deep}. Following their experimental setup, data from 36 of the 45 available infants were used for model development and evaluation, while data from the remaining nine infants were reserved exclusively for hyperparameter optimisation. The selected 36 infants contributed a total of 1,346 five-second motion segments, including 754 FM+ and 592 FM- clips. In each fold, data from 32 infants were used for training and data from the remaining four infants were used for testing. The training set was further divided into training and validation subsets using an 87.5\%/12.5\% split. To strengthen the evaluation, all models were trained with 3 fixed random seeds (42, 67 and 168) and the averaged metrics were reported in Tables~\ref{tab:allRes} and \ref{tab:clientRes}. 

To investigate the effects of data imbalance, the training and testing subjects in each fold were further partitioned across three clients according to subject identity. Three client-distribution settings were evaluated. We denote each setting as \textit{Train-Test}, where the numbers before and after the hyphen represent the relative subject allocation ratios for Clients \#1, \#2, and \#3 in the training and testing sets, respectively. For example, \textit{111-211} denotes a 1:1:1 training split and a 2:1:1 testing split across the three clients. The evaluated settings comprised a balanced-training configuration (\textit{111-211}), a consistently imbalanced configuration (\textit{211-211}), and a distribution-shift configuration (\textit{211-112}). The partitioning was performed at the subject level, such that all motion segments from a given infant were allocated to a single client.

%To simulate a federated learning environment, the training and testing data in each fold were further partitioned across three clients according to subject identity. We further evaluated the performance under 3 different data splits between the 3 clients: with the ratio of 1:1:1 in training and 2:1:1 in testing (111-211), 2:1:1 in both training and testing (211-211), and 2:1:1 in training and 1:1:2 in testing (211-112). Note that the ratios are computed at the subject-level and this partitioning ensured that all motion segments from a given infant were assigned to a single client.%, thereby preserving the subject-level separation required for cross-subject evaluation and mimicking the data silos commonly encountered in multi-centre clinical settings.

%Specifically, the training data were distributed across the three clients using data from \{11, 11, 10\} infants, while the testing data were distributed using data from \{2, 1, 1\} infants, respectively. This partitioning ensured that all motion segments from a given infant were assigned to a single client, thereby preserving the subject-level separation required for cross-subject evaluation and mimicking the data silos commonly encountered in multi-centre clinical settings. 

For automated GMA classification, we adopted the convolutional neural network (CNN) architecture proposed by Kulvicius et al.~\cite{kulvicius2025deep}. The original implementation was provided in Keras with TensorFlow; however, to facilitate integration with the federated learning framework and associated uncertainty-estimation modules, the model was reimplemented in PyTorch while preserving the original network architecture and training procedure. The hyperparameter configuration corresponding to the best-performing model reported in~\cite{kulvicius2025deep} was used throughout all experiments. Readers are referred to the original publication for a detailed description of the network architecture and hyperparameter selection procedure.

\subsection{Implementation details}
%For the federated learning models, model selection is performed at the server level based on the aggregated validation loss across all clients, and the corresponding global model checkpoint is used for inference. 
%During training, the settings include communication rounds $R=100$, local epochs $E=2$, dropout rate = 0.3, 50 and 100 Monte Carlo passes for training and inference, respectively. For the centralized and local training, we followed Kulvicius et al.~\cite{kulvicius2025deep} by picking the checkpoint with the lowest validation loss during training for inference. We also followed their settings with batch size = 4 and learning rate = 0.001 when training the CNN model.

Unless otherwise stated, all experiments used $R=100$ communication rounds, $E=2$ local epochs, using the Adam optimizer with a learning rate of 0.001, a batch size of 4, and a dropout rate of 0.3. MC-Dropout employed 50 stochastic forward passes during training and 100 during inference. Following Kulvicius et al.~\cite{kulvicius2025deep}, the checkpoint with the lowest validation loss was selected for inference in the \textit{Centralized} and \textit{No Federation} settings.

\section{Results} \label{sec:res}
This section presents the quantitative evaluation of the proposed federated learning framework. We first provide an overall comparison by combining the testing results obtained from all three clients. Subsequently, the performance of the individual clients is analyzed to better understand the impact of federated learning and uncertainty-aware aggregation under different local data distributions. 

% Table~\ref{tab:allRes} summarises the performance of the investigated methods using the combined testing results from all clients. 

 \subsubsection{Aggregated Evaluation}
 The aggregated evaluation summarized in Table~\ref{tab:allRes} enables direct comparison with the \textit{Centralized} setting, where all training data are assumed to be available at a single site and a model is trained conventionally without data distribution constraints. As expected, the \textit{No Federation} setting, in which three independent local models were trained without any federated communication, achieved the lowest overall performance across all data splits. %This result is likely attributable to the limited amount of training data available at each client, which restricts the generalisability of the local models and leads to elevated epistemic uncertainty. 
 In contrast, all federated learning approaches, namely \textit{FedAvg} and \textit{UA-FedAvg} (Uncertainty-Aware FedAvg w/ or wo/ vLoss), substantially improved classification performance (over 7\% in accuracy). In their best configurations, the federated models approached the performance of the centralized baseline, with less than a 1\% difference in accuracy.
 
 %The \textit{Local} and \textit{Global} model settings refer to the strategy used for model selection. Under the \textit{Local} setting, each client performs model selection independently using its own validation loss and retains its best local checkpoint for inference. Under the \textit{Global} setting, model selection is performed at the server level based on the aggregated validation loss across all clients, and the corresponding global model checkpoint is used for inference. Notably, the globally selected \textit{UA-FedAvg} model achieved the best overall classification performance, outperforming both \textit{FedAvg} and the centralized baseline across all reported classification metrics. 
 
 %When local model selection was used, \textit{FedAvg} achieved the highest performance in three out of the four classification metrics. This observation suggests that while conventional FedAvg may provide strong local optimisation behaviour, incorporating uncertainty information into the aggregation process can lead to a more effective globally generalizable model. Interestingly, \textit{FedAvg} consistently produced the lowest mean and median epistemic uncertainty under both local and global model-selection settings, indicating a higher degree of confidence in its predictions. Nevertheless, the difference in uncertainty between \textit{FedAvg} and \textit{UA-FedAvg} became substantially smaller when the globally selected models were used, suggesting that global model selection contributes to improved model stability and confidence regardless of the aggregation strategy.

 Of the 3 data splits, \textit{UA-FedAvg w/ vLoss} achieved the highest accuracy, sensitivity, and F1 score in the 111-211 and 211-211 splits, as well as achieving a comparable classification performance in accuracy with \textit{Centralized}. On the other hand, \textit{UA-FedAvg} performed the best in the 211-112 split by topping all 4 classification metrics on such a more challenging and unbalanced split. In terms of predictive uncertainty, \textit{No Federation} consistently produced the lowest mean and standard deviation, whereas \textit{UA-FedAvg} achieved the lowest median predictive uncertainty across the three data splits. 
 
 In summary, while the observed performance gains over FedAvg were modest, the results indicate that incorporating predictive uncertainty into the aggregation process can provide consistent benefits under certain data partitioning schemes.

 \subsubsection{Client-Level Evaluation} 
 Table~\ref{tab:clientRes} summarizes the performance achieved by individual clients under the different data splits. Overall, \textit{UA-FedAvg w/ vLoss} appeared to provide greater benefit for clients with comparatively lower train-to-test ratios, such as Client \#1 in the 111-211 split and Client \#3 in the 211-112 split, both corresponding to a 1:2 ratio. Across the evaluated data splits, \textit{UA-FedAvg} generally achieved higher Accuracy and F1 score than \textit{FedAvg}, with the exception of Client \#1 in the 211-211 split, where the client possessed a relatively larger proportion of the training data. 
 
 With regard to predictive uncertainty, \textit{UA-FedAvg} generally yielded lower median predictive uncertainty across clients, whereas the \textit{No Federation} setting achieved the lowest average predictive uncertainty and standard deviation for most client configurations. These observations suggest that uncertainty-aware aggregation affects the distribution of model confidence across clients, although further investigation is required to better understand the relationship between predictive uncertainty and model reliability in federated settings.
 
 %For these clients, \textit{UA-FedAvg} consistently outperformed \textit{FedAvg} across the reported classification metrics, achieving improvements in F1 Score ranging from 2.21\% to 3.97\%. In contrast, when local model selection was employed, \textit{FedAvg} achieved better performance for Clients \#2 and \#3, indicating that conventional federated averaging may provide stronger local optimisation behaviour in certain client-specific settings. 
 
 %With respect to uncertainty estimation, \textit{FedAvg} consistently produced lower mean and median epistemic uncertainty than \textit{UA-FedAvg} across all clients under both local and global model-selection strategies. This suggests that the predictions generated by \textit{FedAvg} were generally associated with higher confidence. Nevertheless, the differences in epistemic uncertainty between the two aggregation methods were relatively small at the client level and became even less pronounced when globally selected models were used. Taken together, these findings suggest that while uncertainty-aware aggregation can improve classification performance for certain clients, its impact on the predictive uncertainty is modest when compared with standard federated averaging.

\begin{table*} [htb]
\begin{center}
\caption{Classification performance and predictive uncertainty on the combined results with different data splits.} \label{tab:allRes}
\begin{tabular}{ c c c c c c c c c }
\hline
\multirow{ 2}{*}{Method} & \multirow{ 2}{*}{Data Split} & \multirow{ 2}{*}{Accuracy$\uparrow$} & \multirow{ 2}{*}{Sensitivity$\uparrow$} & \multirow{ 2}{*}{Specificity$\uparrow$} & \multirow{ 2}{*}{F1 Score$\uparrow$} & \multicolumn{3}{c}{Predictive Uncertainty$\downarrow$} \\
& & & & & & Average & Median & Std. Dev \\
\hline
 \hline
 Centralized & -- & 0.8685 & 0.8855 & 0.8468 & 0.8830 & 0.0668 & 0.0578 & 0.0553\\
 \hline
 No Federation & \multirow{ 4}{*}{111-211} & 0.7818 & 0.8462 & 0.6999 & 0.8132 & \textbf{0.0826} & 0.0745 & \textbf{0.0640}\\
 FedAvg &  & 0.8531 & \textbf{0.8859} & 0.8114 & 0.8711 & 0.0938 & 0.0615 & 0.0942\\
UA-FedAvg &  & 0.8626 & 0.8828 & \textbf{0.8367} & 0.8781 & 0.0913 & \textbf{0.0596} & 0.0910\\
UA-FedAvg w/ vLoss &  & \textbf{0.8638} & 0.8851 & \textbf{0.8367} & \textbf{0.8792} & 0.0916 & 0.0637 & 0.0865\\
\hline
No Federation & \multirow{ 4}{*}{211-211} & 0.7860 & 0.8351 & 0.7235 & 0.8140 & \textbf{0.0889} & 0.0803 & \textbf{0.0715}\\
FedAvg &  & 0.8564 & 0.8921 & 0.8108 & 0.8743 & 0.0965 & 0.0786 & 0.0845\\
UA-FedAvg &  & 0.8502 & 0.8886 & 0.8012 & 0.8693 & 0.0919 & \textbf{0.0729} & 0.0825\\
UA-FedAvg w/ vLoss & & \textbf{0.8626} & \textbf{0.8970} & \textbf{0.8187} & \textbf{0.8797} & 0.0980 & 0.0758 & 0.0901\\
\hline
No Federation & \multirow{ 4}{*}{211-112} & 0.7717 & 0.7573 & 0.7900 & 0.7880 & \textbf{0.0901} & 0.0826 & \textbf{0.0714}\\
FedAvg & & 0.8566 & 0.8930 & \textbf{0.8102} & 0.8746 & 0.1015 & 0.0741 & 0.0975\\
UA-FedAvg & & \textbf{0.8603} & \textbf{0.9058} & 0.8024 & \textbf{0.8790} & 0.0956 & \textbf{0.0722} & 0.0876\\
UA-FedAvg w/ vLoss & & 0.8549 & 0.8983 & 0.7995 & 0.8739 & 0.1006 & 0.0764 & 0.0915\\
\hline
 \hline
\end{tabular}
\end{center}
\end{table*}

\begin{table*} [htb]
\begin{center}
\caption{Classification performance and predictive uncertainty for each client on different dataset splits.} \label{tab:clientRes}
\begin{tabular}{ c c c c c c c c c c }
\hline
\multirow{ 2}{*}{Data Split} & \multirow{ 2}{*}{Client} & \multirow{ 2}{*}{Method} &  \multirow{ 2}{*}{Accuracy$\uparrow$} & \multirow{ 2}{*}{Sensitivity$\uparrow$} & \multirow{ 2}{*}{Specificity$\uparrow$} & \multirow{ 2}{*}{F1 Score$\uparrow$} & \multicolumn{3}{c}{Predictive Uncertainty$\downarrow$} \\
& & & & & & & Average & Median & Std. Dev \\
\hline
 \hline
 \multirow{ 12}{*}{111-211} &  \multirow{ 4}{*}{\#1} & No Federation & 0.6976 & 0.8553 & 0.5823 & 0.7060 & \textbf{0.0860} & 0.0781 & \textbf{0.0682}\\
& & FedAvg & 0.8484 & 0.9029 & 0.8086 & 0.8345 & 0.0905 & \textbf{0.0525} & 0.0962\\
& & UA-FedAvg & 0.8647 & 0.9011 & 0.8380 & 0.8494 & 0.0874 & 0.0571 & 0.0875\\
& & UA-FedAvg w/ vLoss & \textbf{0.8685} & \textbf{0.9048} & \textbf{0.8420} & \textbf{0.8541} & 0.0865 & 0.0538 & 0.0864\\
\cline{2-10}
& \multirow{ 4}{*}{\#2} & No Federation & 0.8245 & 0.8236 & 0.8268 & 0.8726 & \textbf{0.0776} & 0.0697 & \textbf{0.0581}\\
& & FedAvg & 0.8729 & 0.8706 & 0.8794 & 0.9091 & 0.0908 & 0.0571 & 0.0933\\
& & UA-FedAvg & \textbf{0.8747} & 0.8665 & \textbf{0.8969} & 0.9099 & 0.0887 & \textbf{0.0540} & 0.0905\\
& & UA-FedAvg w/ vLoss & \textbf{0.8747} & \textbf{0.8738} & 0.8772 & \textbf{0.9105} & 0.0902 & 0.0631 & 0.0840\\
\cline{2-10}
& \multirow{ 4}{*}{\#3} & No Federation & 0.8167 & 0.8938 & 0.7522 & 0.8166 & \textbf{0.0865} & 0.0769 & \textbf{0.0664}\\
& & FedAvg & 0.8272 & \textbf{0.9062} & 0.7609 & 0.8270 & 0.1026 & 0.0799 & 0.0919\\
& & UA-FedAvg & \textbf{0.8405} & 0.9042 & 0.7871 & \textbf{0.8383} & 0.1004 & \textbf{0.0742} & 0.0949\\
& & UA-FedAvg w/ vLoss & \textbf{0.8405} & 0.8917 & \textbf{0.7976} & 0.8359 & 0.1001 & 0.0804 & 0.0893\\
\hline
\hline
\multirow{ 12}{*}{211-211} &  \multirow{ 4}{*}{\#1} & No Federation & 0.8310 & \textbf{0.8997} & 0.6447 & 0.8859 & \textbf{0.0771} & \textbf{0.0631} & \textbf{0.0725}\\
& & FedAvg & \textbf{0.8800} & 0.8859 & \textbf{0.8640} & \textbf{0.9151} & 0.0937 & 0.0715 & 0.0847\\
& & UA-FedAvg & 0.8611 & 0.8746 & 0.8246 & 0.9019 & 0.0907 & 0.0715 & 0.0830\\
& & UA-FedAvg w/ vLoss & 0.8735 & 0.8819 & 0.8509 & 0.9106 & 0.0948 & 0.0678 & 0.0904\\
\cline{2-10}
 &  \multirow{ 4}{*}{\#2} &No Federation & 0.7521 & 0.8000 & 0.7120 & 0.7464 & 0.1048 & 0.0968 & \textbf{0.0659}\\
& & FedAvg & 0.8291 & 0.9271 & 0.7469 & 0.8316 & 0.1067 & 0.1016 & 0.0832\\
& & UA-FedAvg & 0.8319 & 0.9292 & 0.7504 & 0.8349 & \textbf{0.1022} & \textbf{0.0869} & 0.0843\\
& & UA-FedAvg w/ vLoss & \textbf{0.8376} & \textbf{0.9313} & \textbf{0.7592} & \textbf{0.8397} & \textbf{0.1022} & 0.0921 & 0.0824\\
\cline{2-10}
 &  \multirow{ 4}{*}{\#3}&No Federation & 0.7548 & 0.7198 & 0.7805 & 0.7136 & 0.0914 & 0.0827 & \textbf{0.0711}\\
& & FedAvg & 0.8476 & 0.8755 & 0.8273 & 0.8297 & 0.0917 & 0.0689 & 0.0836\\
& & UA-FedAvg & 0.8507 & 0.8846 & 0.8260 & 0.8337 & \textbf{0.0851} & \textbf{0.0643} & 0.0792\\
& & UA-FedAvg w/ vLoss & \textbf{0.8685} & \textbf{0.9011} & \textbf{0.8447} & \textbf{0.8526} & 0.0986 & 0.0697 & 0.0948\\
\hline
\hline
\multirow{ 12}{*}{211-112}  &  \multirow{ 4}{*}{\#1} & No Federation & 0.8434 & 0.9136 & 0.7126 & 0.8835 & \textbf{0.0796} & 0.0645 & \textbf{0.0763}\\
& & FedAvg & 0.9147 & 0.9012 & \textbf{0.9397} & 0.9322 & 0.0841 & \textbf{0.0436} & 0.0930\\
& & UA-FedAvg & \textbf{0.9237} & \textbf{0.9336} & 0.9052 & \textbf{0.9409} & 0.0812 & 0.0475 & 0.0850\\
& & UA-FedAvg w/ vLoss & 0.9147 & 0.9228 & 0.8994 & 0.9336 & 0.0843 & 0.0500 & 0.0861\\
\cline{2-10}
 &  \multirow{ 4}{*}{\#2} & No Federation & 0.6250 & 0.6395 & 0.5463 & 0.7415 & 0.1056 & 0.0992 & \textbf{0.0587}\\
& & FedAvg & 0.8477 & 0.8707 & 0.7222 & 0.9062 & 0.1089 & 0.0844 & 0.0981\\
& & UA-FedAvg & \textbf{0.8534} & \textbf{0.8759} & 0.7315 & \textbf{0.9098} & \textbf{0.1049} & \textbf{0.0808} & 0.0928\\
& & UA-FedAvg w/ vLoss & 0.8190 & 0.8333 & \textbf{0.7407} & 0.8858 & 0.1180 & 0.1012 & 0.0943\\
\cline{2-10}
 &  \multirow{ 4}{*}{\#3} & No Federation & 0.7847 & 0.7261 & 0.8303 & 0.7471 & \textbf{0.0899} & 0.0814 & \textbf{0.0717}\\
& & FedAvg & 0.8346 & 0.9006 & \textbf{0.7833} & 0.8265 & 0.1067 & 0.0835 & 0.0983\\
& & UA-FedAvg & 0.8355 & 0.9055 & 0.7811 & 0.8280 & 0.0989 & 0.0810 & 0.0863\\
& & UA-FedAvg w/ vLoss & \textbf{0.8402} & \textbf{0.9201} & 0.7780 & \textbf{0.8343} & 0.1023 & \textbf{0.0790} & 0.0915\\
\hline
 \hline
\end{tabular}
\end{center}
\end{table*}

\section{Conclusion and Future Work}
%In this paper, we presented, to the best of our knowledge, the first federated learning framework for infant movement analysis. As a clinically relevant use case, the proposed framework was evaluated on automated General Movement Assessment (GMA). To improve model interpretability and trustworthiness, Monte Carlo (MC) Dropout was incorporated to estimate epistemic uncertainty, providing an indication of model confidence during inference. Furthermore, we proposed an Uncertainty-Aware Federated Averaging approach (UA-FedAvg), which incorporates predictive entropy derived from MC-Dropout into the aggregation process to improve the robustness and reliability of the global model. Experimental results demonstrated that federated learning substantially improved classification performance compared with independently trained local models, while achieving performance comparable to centralized training. In particular, UA-FedAvg achieved the best overall classification performance when the globally selected model was used, outperforming conventional FedAvg across all evaluated classification metrics. %These findings suggest that incorporating uncertainty information into the aggregation process can improve global model generalisation without requiring access to raw patient data.

In this paper, we presented, to the best of our knowledge, the first federated learning framework for infant movement analysis. As a clinically relevant use case, the proposed framework was evaluated on automated General Movement Assessment (GMA). To quantify model confidence and provide a measure of model confidence, Monte Carlo (MC) Dropout was incorporated to estimate predictive uncertainty during inference. Furthermore, we proposed UA-FedAvg, an uncertainty-aware federated aggregation strategy that incorporates predictive entropy derived from MC-Dropout into the aggregation process. Experimental results demonstrated that federated learning substantially improved classification performance compared with independently trained local models, while achieving performance approaching that of centralized training. Across the evaluated data splits, UA-FedAvg and its variant (i.e. w/ vLoss) generally achieved competitive performance and, in most settings, improved upon conventional FedAvg. These findings suggest that predictive entropy may provide useful information for guiding federated aggregation.

While this study focuses on automated General Movement Assessment, the proposed UA-FedAvg strategy is model-agnostic and only requires a client-level uncertainty estimate. Consequently, it can potentially be extended to other federated learning applications where predictive uncertainty can be estimated and used to guide aggregation.

Although this study demonstrated an interesting direction to work towards using federated learning in handling sensitive data, other considerations such as Information Governance, Computing and Secure Technologies, and Data Management will be needed~\cite{TREvolution}. A limitation of this study is that the dataset comprises recordings from typically developing infants. Consequently, the reported results may not generalize to clinical populations at elevated risk of neurodevelopmental disorders, including infants later diagnosed with CP. Future work will evaluate the proposed framework on more diverse cohorts collected across multiple institutions.

The proposed aggregation strategy assumes that lower predictive entropy is associated with more reliable client predictions. While predictive entropy provides a practical confidence proxy, a low-entropy model may still be overconfident and incorrect under distribution shift. Future work will investigate calibration-based measures (e.g., ECE and reliability diagrams) and alternative uncertainty metrics such as BALD mutual information to better characterize client reliability, and incorporate sample-level information, such as noise and label-flip detection results~\cite{Bradley:FLTA2026}, to inform the aggregation strategy. 
%Several technical directions for future work remain. First, the current uncertainty-aware aggregation scheme employs a relatively simple weighting strategy based on predictive entropy and validation loss. Alternative formulations for combining uncertainty and performance information could be explored to further improve aggregation robustness. Second, the hyperparameter configuration was adopted directly from the centralized CNN model proposed by Kulvicius et al.~\cite{kulvicius2025deep}. While this ensures a fair comparison with prior work, the resulting configuration may not be optimal for the individual client datasets under a federated learning setting. %Future studies will therefore investigate client-specific and federated hyperparameter optimisation strategies. 
Finally, we plan to extend the work by comparing UA-FedAvg to other widely used methods, such as FedProx, SCAFFOLD, FedAdam, and FedYogi, to further evaluate the effectiveness of different approaches with more rigorous statistical analysis.
%. the current federated setup distributes data relatively evenly across the three clients. %Although this allows controlled evaluation of the proposed framework, real-world multi-centre deployments often exhibit substantial imbalances in both dataset size and class distribution. 
%Future work will investigate the performance of the proposed uncertainty-aware aggregation strategy under increasingly heterogeneous and unbalanced client distributions, providing a more realistic assessment of its applicability in clinical environments.

\section*{Acknowledgement}
This work was supported by the UK Research and Innovation DARE UK Real-world Research Exemplar Programme (Ref:~UKRI4079). This study utilized the publicly available infant movement dataset released by Kulvicius et al.~\cite{kulvicius2025deep}.

\bibliographystyle{IEEEtran}
\bibliography{main}

\end{document}